\pdfoutput=1

\documentclass[11pt]{article}

\usepackage[final]{acl}

\usepackage{times}
\usepackage{latexsym}

\usepackage[T1]{fontenc}
\usepackage[utf8]{inputenc}

\usepackage{microtype}

\usepackage{inconsolata}

\usepackage{graphicx}

\usepackage{amsfonts}
\usepackage{amsmath}
\usepackage{algorithm}
\usepackage{algorithmic}
\usepackage[capitalise]{cleveref}
\usepackage{subfig}
\usepackage{booktabs}
\usepackage{multirow}
\usepackage{threeparttable}
\makeatletter
\def\thanks#1{\protected@xdef\@thanks{\@thanks
        \protect\footnotetext{#1}}}
\makeatother

\title{FlexQuant: A Flexible and Efficient Dynamic Precision Switching Framework for LLM Quantization}

\author{
 \textbf{Fangxin Liu\textsuperscript{$1,2,\dagger$}},
 \textbf{Zongwu Wang\textsuperscript{$1,2,\dagger$}},
 \textbf{Jinhong Xia\textsuperscript{3}},
\\
 \textbf{Junping Zhao\textsuperscript{$4,*$}},
 \textbf{Shouren Zhao\textsuperscript{4}},
 \textbf{Jinjin Li\textsuperscript{4}},
 \textbf{Jian Liu\textsuperscript{5}},
\\
 \textbf{Li Jiang\textsuperscript{1,2}},
 \textbf{Haibing Guan\textsuperscript{1}}
\\
\\
 \textsuperscript{1}School of Computer Science, Shanghai Jiao Tong University,
 \textsuperscript{2}Shanghai Qi Zhi Institute,
 \\
 \textsuperscript{3}New York University Shanghai,
 \textsuperscript{4}Ant Group,
 \textsuperscript{5}Beihang University
\\
 \small{
   \textbf{Correspondence:} \href{mailto:liufangxin@sjtu.edu.cn,wangzongwu@sjtu.edu.cn,junping.zjp@antgroup.com}{\{liufangxin,wangzongwu\}@sjtu.edu.cn,junping.zjp@antgroup.com}
 }
 \thanks{This work was done when Jinhong Xia was an intern at Shanghai Jiao Tong University. Junping Zhao and Li Jiang are the corresponding authors.}
}

\begin{document}
\maketitle
\begin{abstract}
The rapid advancement of large language models (LLMs) has exacerbated the memory bottleneck due to the widening gap between model parameter scaling and hardware capabilities. While post-training quantization techniques effectively reduce memory overhead, existing methods predominantly rely on static quantization strategies, which struggle to adapt to dynamic workloads. To address this, we propose FlexQuant, a dynamic precision-switching framework that optimizes the trade-off between inference speed and accuracy. Leveraging model perplexity entropy and Kullback-Leibler divergence, FlexQuant enables fine-grained, layer-wise mixed-precision quantization and dynamically adjusts bit-widths during each token generation. FlexQuant provides a comprehensive analysis of quantization strategies, introduces a precision requirement model for optimal switching, and implements efficient fine-grained precision management. Evaluations demonstrate that FlexQuant achieves a 1.3× end-to-end speedup across diverse language tasks with negligible accuracy loss introduced. This framework offers a flexible and adaptive solution for efficient LLM deployment. Code is released at \url{https://github.com/ZongwuWang/FlexQuant.git}. \end{abstract}

\section{Introduction}

Over the past decade, artificial neural networks (ANNs), particularly large language models (LLMs) based on transformer architectures, have witnessed remarkable advancements, sparking profound transformations across various societal sectors. Yet, these impressive strides in model capabilities have been achieved alongside an exponential growth in model parameters, which have been expanding at a rate of 240-fold every two years. This rapid escalation far outpaces the developmental trajectory of memory hardware, which has been progressing at a comparatively modest rate of 3.1-fold biennially. 
This pronounced disparity between model parameter requirements and hardware capabilities presents a significant bottleneck, constraining the pace at which artificial intelligence can evolve~\cite{liu2024spark}.

Model quantization emerges as a promising technique to alleviate the memory demands, with a plethora of low bit-width quantization algorithms continuously being proposed~\cite{liu2021improving}. In general, model quantization methodologies fall into two primary categories: Quantization-Aware Training (QAT) and Post-Training Quantization (PTQ)~\cite{yang2024searchq}. Given the substantial overhead associated with training LLMs, PTQ algorithms are predominantly employed to facilitate quantizing these models. 
PTQ based on uniform quantization provides an effective approach for compressing full-precision or half-precision floating-point data into lower bit-width integer representations, such as 8-bit or fewer, thereby significantly reducing memory overhead. To further enhance quantization performance in LLMs, advanced techniques have been developed to address the challenges posed by anomalous values. Activation-aware weight quantization optimizes the quantization process by considering the distribution of activations\cite{lin2024awq}, while outlier-aware mixed-precision quantization dynamically adjusts bit-width allocation to better handle outliers\cite{dettmers2022gpt3}. Additionally, the incorporation of approximate second-order information improves quantization accuracy by capturing more precise gradient statistics\cite{frantar2022gptq}. Clustering-based non-uniform quantization offers another refinement by grouping similar values and applying tailored quantization levels, thereby mitigating the impact of irregular value distributions\cite{zhang2024kv}. 

These prevailing state-of-the-art quantization strategies demonstrate varying degrees of precision and performance efficiency of LLM deployment. However, they often resort to a single, invariant quantization strategy throughout model inference, notwithstanding the dynamic nature of token generations. Consequently, no single quantization model consistently attains optimal accuracy and speed across diverse workloads.

To navigate this challenge, there arises a necessity to judiciously balance latency and precision across different inference stages. This requires a mechanism to dynamically switch between quantization strategies, thereby achieving peak inference speed with negligible accuracy loss. Such a solution entails addressing the following three challenges:
\textbf{(1)} Analyzing the precision and performance of various quantization strategies;
\textbf{(2)} Determining precision switching points during inference, necessitating robust accuracy requirement modeling;
\textbf{(3)} Efficient implementation of fine-grained precision management.

Considering the above challenges, the progressive mixed-precision decoding (PMPD) \cite{chen2024progressive} proposes a gradual decrease in precision needs as token generation advances during decoding.
PMPD entails a gradual reduction in the bit-width of weight quantization to strike a balance between accuracy and efficiency demands. The method introduces two precision switching schemes: a prompt-agnostic static scheduler and a task-agnostic learned scheduler. However, it lacks a robust theoretical analysis of precision requirements, leading to heuristic-driven precision switching schedules that fall short of optimal performance. Furthermore, the Any-Precision LLM\cite{park2024any}-based quantization scheme is unable to deliver end-to-end performance enhancements in GPU systems because of inefficient memory access patterns.

In view of the defects of PMPD in precision management and memory access, in this paper, we propose a token-wise quantization precision switching framework, named FlexQuant. FlexQuant models precision demands during the decoding phase with fine granularity by leveraging model perplexity entropy, facilitating optimal dynamic precision switching. Additionally, by integrating layer-wise mixed precision switching based on Kullback–Leibler (KL) divergence, FlexQuant offers fine-grained management of the model's effective quantization bit-width, striking an optimal balance between precision and performance. The main contributions of this work can be summarized as follows:

\begin{itemize}
\item \textbf{Precision Requirement Analysis and Management:} 
A comprehensive analysis of the precision requirement span the sequence inference lifetime. Based on this, perplexity entropy-based light-weight scheduler is proposed for efficient precision management.
\item \textbf{Fine-Grained Precision Management:} A KL divergence–guided layer-wise mixed-precision scheme is proposed, enabling continuous precision switching while optimizing memory access efficiency. 
\item \textbf{Dynamic Precision Switching Framework:} A flexible inference framework FlexQuant, which supports token granularity and precision switching. Experiments across diverse tasks show that FlexQuant achieves 1.3x end-to-end speedup over baseline models. 
\end{itemize} 
\section{Background}

\subsection{Autoregressive LLM}

\begin{figure}[t]
\setlength{\abovecaptionskip}{3pt}  %
\setlength{\belowcaptionskip}{3pt} %
\centering
\includegraphics[width=0.95\linewidth]{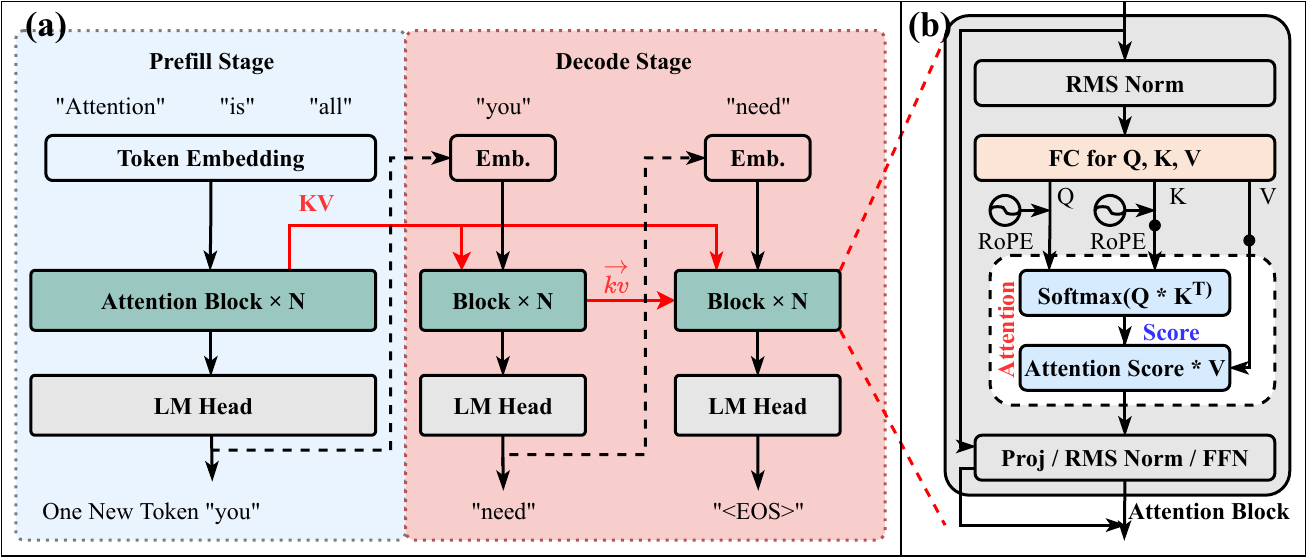}
\caption{Overview of transformer model structure. (a) Transformer model consists of two stage, the prefill stage processes the prompt prefilling of all tokens in batches, and the decode stage generates tokens one by one; (b) Details of the attention block.}
\label{fig:llm}
\vspace{-0.4cm}
\end{figure}

Transformers are the foundation of modern NLP and deep learning. Unlike RNNs and LSTMs, the Transformer model relies on self-attention, enabling parallelization and more efficient training.
The Transformer model architecture consists of two main stages: the Prefill stage and the Decode stage, as depicted in Fig.~\ref{fig:llm}(a). In the Prefill stage, the model processes the entire input sequence in batches, and generate the key and value vectors for each prompt tokens simultaneously.
The Decode stage operates auto-regressively, generating the output sequence token by token. 
The auto-regressively interaction is handled by the attention blocks, which is the most important modules in the transformer models. 
The detailed algorithm of the attention block is depicted in Fig.~\ref{fig:llm}(b). 
The unique attention mechanism, highlighted with a white background, enables token interactions within the context. This process is formalized as follows:
\begin{equation}
\label{eq:attn}
    A_n = \mathrm{Softmax}(\frac{Q \times K^T}{\sqrt{d_k}}) \times V
\end{equation}

Where $Q \in \mathbb{R}^{n_q \times d_k}$, $K$ and $V \in \mathbb{R}^{n_k \times d_k}$ represent the query, key, and value vectors corresponding to each token, $d_k$ denotes the head dimension, $n_q$ and $n_k$ denote the number of query and key/value for the attention computation, respectively. Due to the causality of the language model, the query vector of each token needs to be interacted with all previous key/value vectors. 
Then the attention scores are utilized to calculate the weighted sum of the value vectors for each token, resulting in the current context state vector. This is followed by feature extraction using additional FFN layers.

\subsection{Quantization}

Existing integer quantization maps high-precision numbers to discrete levels, the process can be formulated as:
\begin{equation}
\label{eq:quant}
\left\{
\begin{array}{l}
    \boldsymbol{Q_X} = \lceil \frac{\boldsymbol{X}}{s} + z \rfloor \\
    s = \frac{\boldsymbol{X}_{max} - \boldsymbol{X}_{min}}{q_{max} - q_{min}} \\
    z = \lceil q_{min} - \frac{\boldsymbol{X}_{min}}{s} \rfloor
\end{array}
\right.
\end{equation}
where $\boldsymbol{X}$ is the floating point tensor, $\boldsymbol{Q_X}$ is its n-bit quantized counterpart, $s$ is the scaling factor and $z$ is the zero point. Thus, the dequantized tensor can be represented as,
\begin{equation}
\label{eq:dequant}
    \hat{\boldsymbol{X}} = Q(\boldsymbol{X}) = (\boldsymbol{Q_X} - z) \cdot s
\end{equation}
This is known as asymmetric quantization, where $\boldsymbol{X}{max} = max(\boldsymbol{X})$, $\boldsymbol{X}{min} = min(\boldsymbol{X})$, and $q_{max} - q_{min} = 2^n - 1$ for n-bit integer quantization. Eq.~\ref{eq:quant} can be further simplified to symmetric quantization, where the zero point $z$ is set to 0, and the range is centered around zero with $\boldsymbol{X}{max} = -\boldsymbol{X}{min} = max |\boldsymbol{X}|$. In this case, $q_{max} - q_{min} = 2^n - 2$, as one bit is reserved for the sign.

The de-quantization process of integer quantization needs to be implemented using general-purpose CUDA cores, which are less efficient for this type of operation. This implementation introduces a significant performance bottleneck, resulting in 20\% to 90\% overhead \cite{lin2024qserve}.

\subsection{LLM Perplexity}

Perplexity is a key metric for evaluating the performance of LLMs in NLP tasks, and it is also relevant for precision scheduling in FlexQuant. It measures how well a model predicts a sequence of tokens, with lower perplexity indicating better performance. Mathematically, perplexity is defined as the exponential of the average negative log-likelihood per token:  

\[
\text{Perplexity} = \exp\left(-\frac{1}{N}\sum_{i=1}^{N} \log P(w_i | w_{<i})\right),
\]  

where \(N\) is the total number of tokens, and \(P(w_i | w_{<i})\) is the model's predicted probability for the \(i\)-th token given the preceding tokens. In LLMs, the next-token prediction is based on the logits (unnormalized scores) produced by the final layer of the model. These logits are converted into probabilities via the softmax function:  

\[
P(w_i | w_{<i}) = \text{softmax}(\mathbf{z})_i = \frac{e^{z_i}}{\sum_{j} e^{z_j}},
\]  

where \(\mathbf{z}\) represents the logits for all possible tokens. The model selects the next token either by sampling from this distribution or choosing the token with the highest probability (greedy decoding). By minimizing perplexity during training, the model improves its ability to predict coherent and contextually appropriate sequences.

\section{Motivation}

\subsection{Differences in KL divergence between Layers.}

\begin{figure}[t]
\setlength{\abovecaptionskip}{3pt}  %
\setlength{\belowcaptionskip}{3pt} %
\centering
\includegraphics[width=0.99\linewidth]{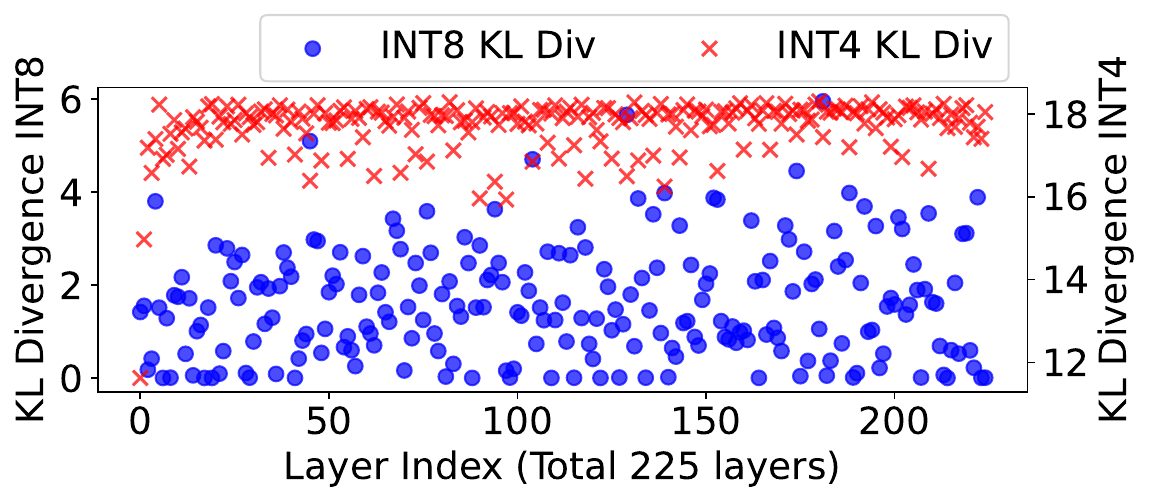}
\caption{The KL divergence of int8 and int4 quantization of the Llama3-8b linear layers.}
\label{fig:kldiv}
\vspace{-0.4cm}
\end{figure}

To investigate the correlation between weight quantization and model accuracy, an experiment was conducted to examine the KL divergence of weight quantization across various linear layers in the Llama3-8B model. The KL divergence in the experiment denotes the variance in distribution between the quantized $\hat{W}$ in comparison to the original full-precision weight $W$, with the findings illustrated in \cref{fig:kldiv}. The results indicate that employing low bit-width quantization results in a notable rise in KL divergence. Moreover, there exist substantial disparities in KL divergence among different Linear layers, suggesting that quantization at distinct layers influences model accuracy differently.

\subsection{Gradual decrease in precision requirements during the decoding phase.}
\label{sec:gradual}

\begin{figure}[t]
\setlength{\abovecaptionskip}{3pt}  %
\setlength{\belowcaptionskip}{3pt} %
\centering
\includegraphics[width=0.95\linewidth]{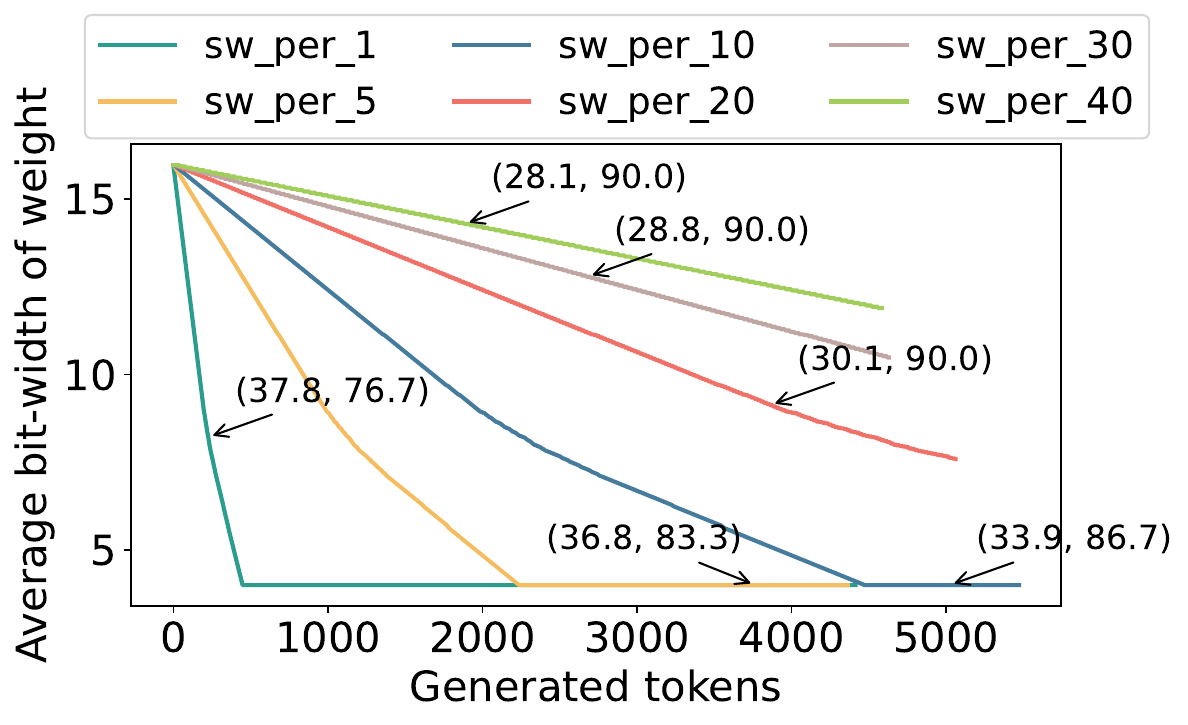}
\caption{The influence of the different bit-width switching speed on performance.}
\label{fig:bitdec}
\vspace{-0.4cm}
\end{figure}

Inspired by \cite{chen2024progressive}, as more and more tokens are generated in the decoding stage, the weight precision of the model can gradually decrease without affecting the accuracy. To explore the impact of the decrease in model precision on performance, we used AutoGPTQ\cite{frantar2022gptq} to quantize the half-floating weight of Llama3-8B model to 8 bits and 4 bits, and then evaluate different precision switching speed with the FlexQuant framework. 
\Cref{fig:bitdec} demonstrates the impact of accuracy switching speed on the inference accuracy of LongWriter dataset. It can be found that when the model precision switching speed is reasonable, it can effectively reduce the memory bandwidth of the model while ensuring that the accuracy is not affected, thus optimizing the generation speed. For example, when switching the bit-width of a linear layer after every 20 tokens is generated will not affect the accuracy of the model, while switching the accuracy every 10 tokens can achieve 20\% of the end-to-end speedup with only 3.3\% accuracy sacrifice.

The result illustrates that an appropriate precision switching speed is crucial to maintain the inference accuracy. Whereas, setting the precision switching speed too fast will result in a decline in inference accuracy. Hence, selecting an optimal precision switching speed is essential to strike a balance between accuracy and performance.

\section{Methodology}

\subsection{Problem Definition}

In long-context generation tasks, the inference cost during the decoding phase is predominantly influenced by the memory bandwidth constraints. To optimize performance during the generation phase, low-precision quantization emerges as an effective strategy. 
During the prefill phase of the model, token parallelism effectively amortizes the latency associated with weight loading, allowing W8A8 quantization to take full advantage of the INT8 computational capabilities of Tensor cores. In contrast, the decoding phase necessitates autoregressive sequential token generation, which requires repetitive loading of model weights, leading to severe memory access bottlenecks. Directly applying lower bit-width quantization, such as INT4, would result in unacceptable accuracy losses.
Thus, we introduce FlexQuant, a lightweight online dynamic quantization precision adjustment inference framework. It aims to balance the trade-offs between model accuracy and efficiency by intelligently adapting the quantization bit-width based on the specific requirements during dynamic token generation. This approach seeks to enhance the overall throughput of generative models while maintaining an acceptable accuracy loss.

\subsection{Precision Requirement Analysis}

In \cref{sec:gradual}, we investigated how appropriately modifying the quantization precision during the decoding phase can enhance the performance of the inference without compromising the accuracy. In this section, we will now explore the root cause behind this observation. 
Firstly, the generation of each token in the decoding phase can be represented as follows:

\begin{equation}
    \label{eq:decoding}
    P(x_i|x_{< i}) = softmax(logits)
\end{equation}

Here, the $logits$ are the outputs of the Transformer model. After being normalized by the softmax function, they represent the probability of each token in the vocabulary being the next token. In deterministic output scenarios, the token with the highest probability is typically selected as the generated token.

\begin{figure}[h]
\setlength{\abovecaptionskip}{3pt}  %
\setlength{\belowcaptionskip}{3pt} %
	\centering
	\subfloat[Fault tolerance of Llama2-7B.\label{fig:l2margin}]{%
    \setlength{\abovecaptionskip}{3pt}  %
    \setlength{\belowcaptionskip}{3pt} %
	\centering
	\begin{minipage}[c]{0.9\linewidth}
	\centering
	\includegraphics[width=\linewidth]{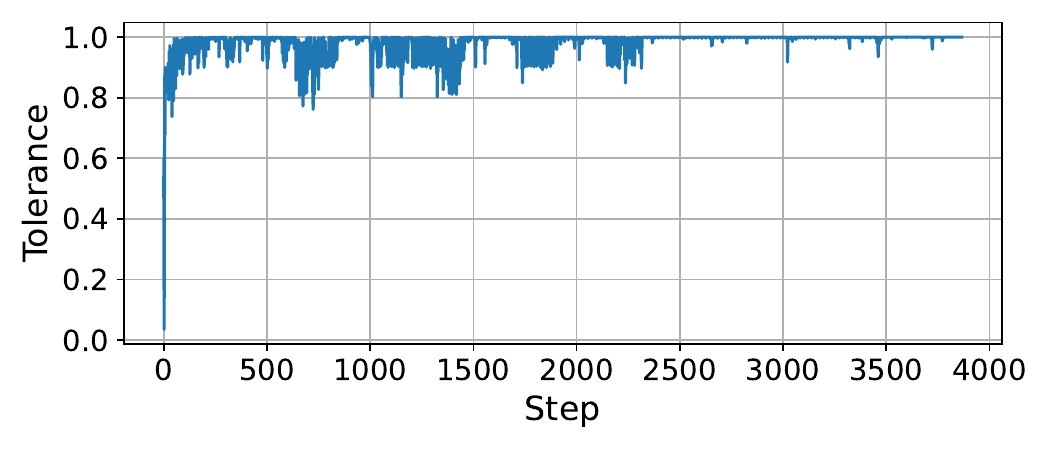}
	\end{minipage}
	}
	\vfill
	\subfloat[Fault tolerance of Llama3-8B.\label{fig:l3margin}]{%
    \setlength{\abovecaptionskip}{3pt}  %
    \setlength{\belowcaptionskip}{3pt} %
	\centering
	\begin{minipage}[c]{0.9\linewidth}
	\centering
	\includegraphics[width=\linewidth]{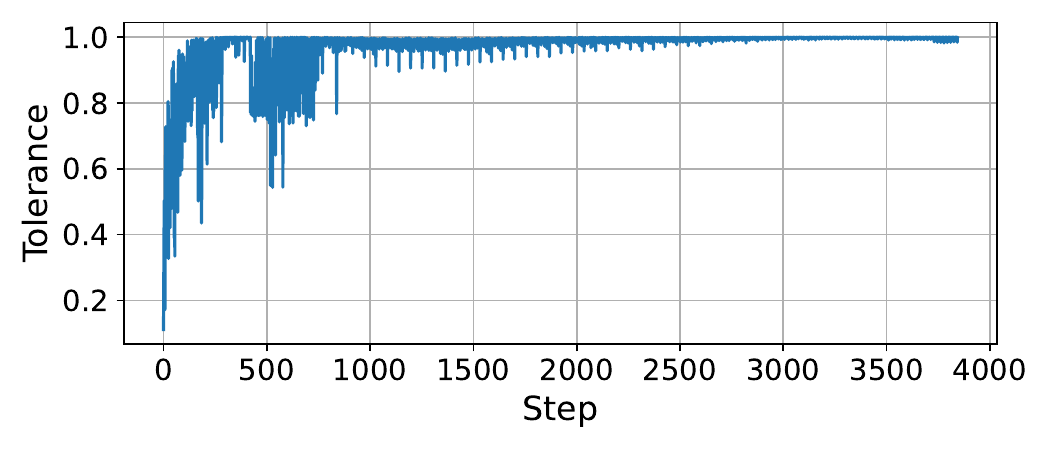}
	\end{minipage}
	}
\caption{Fault tolerance across various models on summarizing CNN/Daily Mail task.}
\label{fig:tol}
\vspace{-0.5cm}
\end{figure}

Therefore, we only need to ensure that the $argmax(P(x_i|x_{< i}))$ output for each token generation remains unchanged. We define the model's fault tolerance as the variance between the maximum probability and the next highest probability, and \cref{fig:tol} illustrates the fault tolerance across various models when summarizing CNN/Daily Mail. It is evident that current models maintain a reasonable level of fault tolerance in the generation process, providing avenues to enhance performance.
As the context length for inference grows, the conditional probability linked to \cref{eq:decoding} will increase gradually, signifying an increase in the model's fault tolerance. Consequently, we can slightly decrease the model's quantization precision to enhance performance without compromising inference accuracy. To accurately quantify the model's fault tolerance at various stages, we introduce the concept of perplexity entropy (PPLE), defined as the exponent of the logits' entropy:
\begin{equation}
    \label{eq:entropy}
    PPLE = exp(-\sum (x_i|x_{< i}) \times log((x_i|x_{< i})))
\end{equation}

\begin{figure}[t]
\setlength{\abovecaptionskip}{3pt}  %
\setlength{\belowcaptionskip}{3pt} %
\centering
\includegraphics[width=0.95\linewidth]{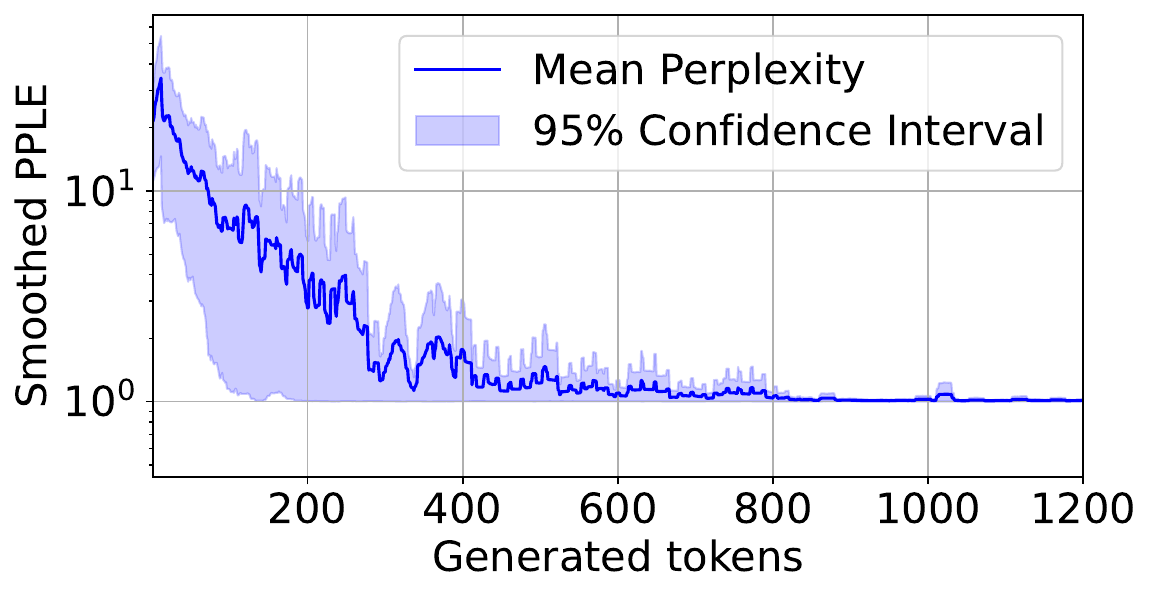}
\caption{Relationship between PPLE and number of generated tokens(smoothed with 20 tokens).}
\label{fig:smpple}
\vspace{-0.2cm}
\end{figure}

\Cref{fig:smpple} demonstrates the correlation between PPLE and the quantity of tokens generated when summarizing the CNN/DM dataset using the llama3-8b model. To accommodate fluctuations in token probabilities, data from 50 samples were gathered and their distributions analyzed. The findings suggest that the mean PPLE across a context serves as a reliable metric for assessing the precision of the model.

\subsection{Mixed-Precision Decoding Scheme} 
In \cite{chen2024progressive}, the any-precision LLM was adopted to achieve progressive precision reduction. However, this approach has limitations. Firstly, it can only lower the bit-width of the entire model parameters synchronously, which results in a coarse-grained precision adjustment that fails to optimize model performance effectively. Additionally, accessing model parameters in arbitrary precision requires bit-level indexing, hindering efficient memory access.
In contrast, FlexQuant introduces the concept of inter-layer mixed precision. Each linear layer's model parameters are stored in multiple formats, such as INT8 and INT4 formats. During the prefill phase, W8A8 quantization is employed, allowing the INT8 parameters to be loaded into the GPU's HBM. 
In the decoding phase, FlexQuant dynamically replaces the INT8 weights of different layers with INT4 weights based on the variations of PPLE. The order of replacement is determined through offline KL divergence analysis, where linear layers with smaller KL divergences will be prioritized for replacement to 4 bits, thus achieving the optimal balance between accuracy and speed.

\subsection{Token-wise Precision Switching Management}
FlexQuant utilizes PPLE to manage the quantization precision of the model in a fine-grained manner. Through extensive experimentation, we discovered that the average PPLE of a continuous sequence of 20 generated tokens can reliably reflect the model's accuracy, providing a stable metric for performance assessment. Consequently, during the decoding phase, we maintain a sliding window of length 20 to compute the moving average of PPLE.
As context continue grows, we observe a consistent decrease in the average PPLE. This decline is indicative of the model's potential efficiency in processing tokens in relation to its output accuracy. When the moving average of PPLE dips below a predefined threshold, it suggests that the model's fault tolerance is excessively high. In this scenario, we can conclude that there is an opportunity to enhance computational efficiency without sacrificing quality.
Subsequently, FlexQuant enables the transition of certain model weights from INT8 to INT4 bit-width. This dynamic adjustment allows the model to optimize its performance by reducing the memory footprint and improving the processing speed without introducing significant accuracy loss. By judiciously applying this switching mechanism based on real-time performance metrics, FlexQuant effectively balances the trade-off between computational efficiency and model fidelity, thereby ensuring that the generative model operates at its best across varying contexts.

\subsection{FlexQuant Implementation}
\begin{algorithm}[tb]
   \caption{FlexQuant Dynamic Precision Inference}
   \label{alg:flexquant}
\begin{algorithmic}
   \STATE {\bfseries Input:} Prompt $P$ of $n$ tokens
   \STATE {\bfseries Output:} Sequence $S$
   \STATE
   $logit_{n-1}$ $\leftarrow$ Prefill($P$) with WxAy model
   \STATE $T_n$ $\leftarrow argmax(logit_{n-1})$
   \STATE $S$ $\leftarrow concat(P, T_n)$
   \textcolor{blue}{\STATE $thre$ $\leftarrow ave(exp(entropy(logits)))$}
   \FOR{$i=0$ {\bfseries to} $M$}
   \STATE $logit_{n+i}$ $\leftarrow$ Decoding($S$)
   \STATE $T_{n+i+1}$  $\leftarrow argmax(logit_{n+i})$
   \STATE $S$ $\leftarrow concat(S, T_{n+i+1})$
   \textcolor{blue}{\STATE $PPLE$ $\leftarrow exp(entropy(logit_{n+i})$ 
   \IF{$slide\_ave(PPLE) < thre$}
   \STATE Replace Wx with lower bit-width of specific- Linear layers
   \ENDIF}
   \ENDFOR
\end{algorithmic}
\end{algorithm}

The inference process for the progressive precision switching in FlexQuant is illustrated in \Cref{alg:flexquant}, and we also highlight the extra operation in blue. Initially, FlexQuant employs W8A8 to prefill the input prompt, effectively leveraging the INT8 computational capabilities of Tensor cores to alleviate the computational bottleneck during this phase. Once the prefill is complete, the first token is generated based on the logits produced by the final model output.
During the decoding phase, FlexQuant updates the sliding average of the PPLE for each token generated based on the corresponding logits. If the sliding average falls below a specified threshold, denoted as \(thre\), part of model weights can be switched from 8-bit quantization to 4-bit quantization. The selection of weights for precision switching is determined through offline KL divergence analysis. In FlexQuant, the threshold \(thre\) is set as the mean PPLE of the last few tokens generated during the prefill phase.
It is important to note that after a precision switch occurs, we must wait until a sufficient number of tokens have been generated before obtaining an effective PPLE for the new precision. Only then can we continue to evaluate whether further adjustments to the quantization precision are necessary. 
\section{Evaluation}

\subsection{Experimental Setup}

\begin{table*}[]
\setlength{\abovecaptionskip}{3pt}  %
\setlength{\belowcaptionskip}{3pt} %
\scriptsize
\begin{threeparttable}
\centering
\caption{Performance comparison of different precision schedulers against low-precision (baselinel) and high-precision (baseline-h) baselines.}
\label{tab:accuracy}
\begin{tabular}{@{}ccccccccccc@{}}
\toprule
\multirow{2}{*}{Model}                                                          & \multirow{2}{*}{Method} & \multicolumn{3}{c}{CNN/DM}  & \multicolumn{3}{c}{Dialogsum} & \multicolumn{3}{c}{IWSLT}  \\ \cmidrule(l){3-11} 
                                                                                &                         & Bits & Rouge-L  & BERTScore & Bits  & Rouge-L   & BERTScore & Bits & Rouge-L & BLEUScore \\ \midrule
\multirow{6}{*}{Vicuna-7B}                                                      & Baseline-l              & 2    & 4.29     & 76.74     & 2     & 1.57      & 73.5      & 2    & 2.24    & 0         \\
                                                                                & Baseline-h              & 3    & 22.27    & 87.01     & 3     & 23.57     & 87.56     & 3    & 52.98   & 25.65     \\
                                                                                & DNS*                    & 2.39 & 22.27    & 86.92299  & 2     & -         & -         & 2.68 & -       & 22.39245  \\
                                                                                & PMPD-Static*            & 2.39 & 22.35908 & 87.09701  & 2     & 24.15925  & 87.56     & 2.68 & -       & 25.2396   \\
                                                                                & PMPD-Learned*           & 2.39 & 22.09184 & 86.83598  & 2.74  & 23.66428  & 87.56     & 2.37 & -       & 24.2649   \\
                                                                                & FlexQuant               & 2.56 & 20.04    & 85.24     & 2.74  & 23.2      & 87.23     & 2.68 & 52.41   & 20.91     \\ \midrule
\multirow{6}{*}{MobileLlaMA}                                                    & Baseline-l              & 3    & 9.3      & 78.93     & 3     & 4.42      & 78.48     & 3    & 6.35    & 0.39      \\
                                                                                & Baseline-h              & 4    & 14.64    & 83.99     & 4     & 10.32     & 81.48     & 4    & 19.89   & 4.51      \\
                                                                                & DNS*                    & 3.37 & 14.81568 & 83.99     & 3.21  & 9.03      & 80.99112  & 3.65 & -       & 4.26195   \\
                                                                                & PMPD-Static*            & 3.37 & 14.97672 & 84.15798  & 3     & 10.44384  & 81.56148  & 3.65 & -       & 4.47392   \\
                                                                                & PMPD-Learned*           & 3.37 & 14.1276  & 83.65404  & 3.21  & 10.50576  & 81.56148  & 3.48 & -       & 4.18979   \\
                                                                                & FlexQuant               & 3.37 & 16.36    & 84.31     & 3.37  & 10.57     & 81.65     & 3.5  & 24.65   & 6.83      \\ \midrule
\multirow{6}{*}{\begin{tabular}[c]{@{}c@{}}Phi-1.5/\\ Zephyr-3B**\end{tabular}} & Baseline-l              & 3    & 7.06     & 77.72     & 3     & 7.41      & 80.79     & 3    & 11.82   & 1.96      \\
                                                                                & Baseline-h              & 4    & 12.27    & 83.31     & 4     & 9.4       & 81.87     & 4    & 55.01   & 24.31     \\
                                                                                & DNS*                    & 3.71 & 9.38655  & 81.14394  & 3.3   & 8.4036    & 81.46065  & 3.34 & -       & 22.55968  \\
                                                                                & PMPD-Static*            & 3.71 & 12.27    & 83.31     & 3.3   & 9.4564    & 81.95187  & 3    & -       & 23.75087  \\
                                                                                & PMPD-Learned*           & 3.09 & 11.74239 & 82.72683  & 3.52  & 9.3436    & 81.78813  & 3.34 & -       & 23.8238   \\
                                                                                & FlexQuant               & 3.11 & 12.73    & 83.18     & 3.52  & 10.86     & 82.86     & 3.44 & 54.58   & 22.88     \\ \bottomrule
\end{tabular}
\begin{tablenotes}
\item[*] Indicates that the data is scaled by the results in PMPD to the baseline-h of this paper.
\item[**] The Phi-1.5 model is used to evaluate the CNN/DM and Dialogsum tasks, and the Zephyr-3B model is used to evaluate the IWSLT task.
\end{tablenotes}
\end{threeparttable}
\vspace{-0.2cm}
\end{table*} 

\begin{table*}[]
\setlength{\abovecaptionskip}{3pt}  %
\setlength{\belowcaptionskip}{3pt} %
\footnotesize
\centering
\caption{Accuracy evaluation of FlexQuant on LongWriter Evaluation Set.}
\label{tab:longwriter}
\begin{tabular}{@{}lllllllll@{}}
\toprule
Bits              & FP16  & INT-8 & INT-4 & FQ-1.3 & FQ-1.5 & FQ-1.7 & FQ-2.0 & FQ-2.5 \\ \midrule
Rouge-L            & 100   & 46    & 26    & 89.43 & 75.44 & 66.58 & 60.86 & 56.14 \\
DeepSeek-V3 Score & 89.02 & 87.63 & 81.66 & 88.54  & 88.19  & 89.23  & 88.33  & 87.6   \\ \bottomrule
\end{tabular}
\vspace{-0.2cm}
\end{table*}

\textbf{Models and Datasets.} 
To ensure a fair comparison with prior work, we adopt the same LLM models and evaluation settings used in PMPD. Specifically, we evaluate edge-friendly models such as Vicuna-7B~\cite{chiang2023vicuna}, MobileLLaMA-1.4B~\cite{chu2023mobilevlm}, Zephyr-3B~\cite{zephyr}, and Phi-1.5~\cite{li2023textbooks}, following PMPD’s zero-shot evaluation protocol. The benchmark includes CNN/DM~\cite{hermann2015teaching} for news summarization, DialogSum~\cite{chen2021dialogsum} for dialogue summarization, and IWSLT Fr-En~\cite{cettolo2017overview} for translation.
In addition, to better evaluate the effect of FlexQuant's asymptotic precision scheduling in long-context generation, we include the LongWriter-LLaMA3.1-8B model on the LongWriter~\cite{bai2024longwriter} benchmark.

\textbf{Baselines}
In this paper, we use the GPTQ algorithm for integer quantization with different accuracies as baselines, and both PMPD and Dense-and-Sparse decomposition(DNS)\cite{kim2023squeezellm} performance data are derived from the publicly available data of the PMPD paper. Since we cannot reproduce the publicly available inference accuracy of the PMPD paper using the public model and dataset, this paper uses the reproduced accuracy of baseline\_h as the baseline and scales the DNS's and PMPD's results according to the accuracy loss percentage compared to PMPD's baseline\_h. Finally, we implement FlexQuant based on huggingface's transformers framework, which only demonstrates the model accuracy by fake quantization since the 2-bit data format is not supported by efficient CUDA kernel for the time being. And in the higher complexity LongWriter scenario, FlexQuant is evaluated for accuracy and speed based on the marlin kernel of int8 and the exllama-v1 kernel of int4. And considering the complexity and quantization sensitivity, half-floating model is set as the switching baseline for this GQA model.

\subsection{Evaluation Results}

\textbf{Model Accuracy.}
We conduct extensive experiments to evaluate the effectiveness of our precision scheduling approach. As shown in \cref{tab:accuracy}, our method achieves similar quantization bit-width and inference performance to DNS and PMPD, while using a simpler and more flexible scheme. However, we observe that these standard classification and generation tasks (e.g., Wikitext, SST-2, and MNLI) may not provide a reliable evaluation of precision scheduling. These tasks are relatively simple, and the resulting small accuracy gaps under different quantization schemes can be easily masked by statistical noise or task insensitivity to quantization perturbations.

To better reflect the value of our precision scheduler, we focus on more challenging long-context generation tasks, where maintaining accuracy under ultra-low-bit quantization is substantially harder. As shown in \Cref{tab:longwriter}, such as the LongWriter task, our method shows clear advantages. For instance, while static INT8 or INT4 quantization causes Rouge-L scores to degrade severely (to 46 and 26 respectively), FlexQuant maintains significantly higher generation quality (e.g., 59 with FQ-2.5), and achieves comparable performance to INT8 models while allowing for more aggressive compression.
Notably, FlexQuant introduces a controllable trade-off between compression and quality via its precision-perplexity threshold. This tunability is absent in static schemes, making FlexQuant better suited for deployment under varying resource budgets.
In addition, we follow LongWriter's evaluation methodology and assess generation quality using DeepSeek-V3. FlexQuant again matches INT8-level performance under stricter evaluation, reinforcing its robustness.
While FlexQuant may not yield higher accuracy on simple tasks, it provides clear, quantifiable benefits in complex scenarios, where static quantization fails to balance quality and efficiency. These results validate the practical utility of our precision scheduler.

\begin{figure}[t]
\setlength{\abovecaptionskip}{3pt}  %
\setlength{\belowcaptionskip}{3pt} %
\centering
\includegraphics[width=0.93\linewidth]{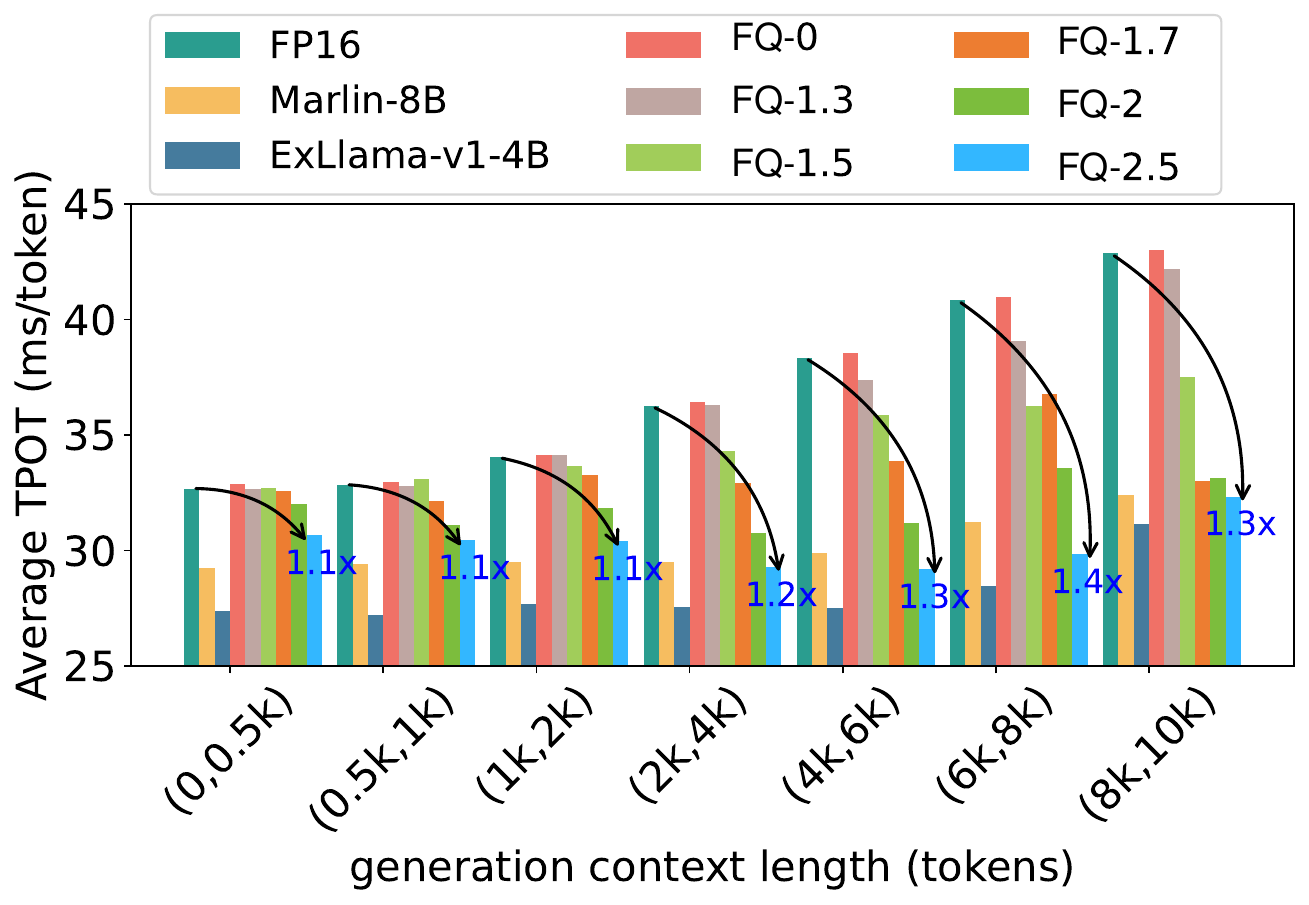}
\caption{End-to-end TPOT evaluation results on GPU.}
\label{fig:speed}
\vspace{-0.4cm}
\end{figure}

\textbf{TPOT Performance.}
We further evaluate FlexQuant’s inference performance on GPUs using the LongWriter task. \Cref{fig:speed} presents the end-to-end generation speed comparison of different quantization schemes on the LongWriter task. As the generation length increases, the decoding speed gradually decreases due to the growing computational and memory access overhead of KV Cache, which aligns with expectations for Transformer-based auto-regressive models.

Compared to static quantization, FlexQuant achieves increasing speedup over FP16 as the sequence grows, eventually reaching 1.3×. This improvement occurs because FlexQuant dynamically transitions model weights from FP16 to INT8 and INT4 formats as generation length increases, thereby optimizing the memory bandwidth bottleneck. However, since our current evaluation doesn't apply quantization to KV Cache, the end-to-end performance gains from weight quantization are gradually offset by the increasing attention latency at longer sequence lengths.

INT8 quantization reduces memory bandwidth consumption by half compared to FP16, providing consistent performance gains across sequence lengths. While the ExLlama-v1 kernel shows limited improvements for INT4 formats due to additional handling overhead, \cref{fig:speed} demonstrates that FQ-2.5 achieves faster generation than INT8 for sequences between 2k and 4k tokens. This highlights the potential of ultra-low-bit quantization, although the overall speedup is currently bounded by the efficiency of the INT4 kernel implementation.

\begin{figure}[t]
\setlength{\abovecaptionskip}{3pt}  %
\setlength{\belowcaptionskip}{3pt} %
\centering
\includegraphics[width=0.93\linewidth]{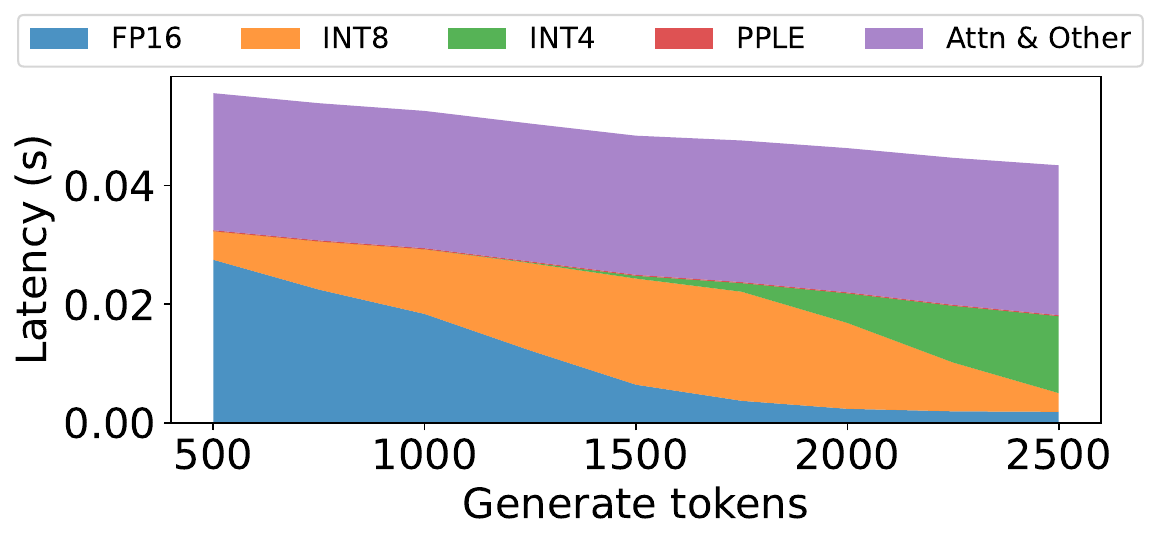}
\caption{Latency Breakdown of FlexQuant on LlamaWriter with thre=2.}
\label{fig:latbrk}
\end{figure}

\textbf{Latency Breakdown.}
Furthermore, we also analyzed the scaling of the latency breakdown as tokens generation advanced, and the results is shown in \Cref{fig:latbrk}. It is evident that FlexQuant gradually shifts the computational overhead from FP16 to INT8, and finally to INT4. Additionally, although the latency of KV Cache increases slowly with the number of tokens, FlexQuant, by optimizing the weight access bottleneck, achieves a decrease in end-to-end latency. The red part in the figure represents the PPLE computation latency introduced by FlexQuant, which can be negligible as it only involves an additional cross-entropy calculation.

\section{Related Works}
\textbf{Weight Quantization.}
Recent advances in weight quantization have pushed precision to extremely low bit-widths (e.g., $\le$4-bit). AWQ preserves important weights by applying channel-wise scaling based on activation statistics~\cite{lin2024awq}. SmoothQuant~\cite{xiao2023smoothquant} alleviates activation quantization difficulty by mathematically shifting it to weights, enabling smoother quantization. QuIP\#\cite{tseng2024quip} and AQLM\cite{egiazarian2024extreme} achieve state-of-the-art performance under 2-3 bit quantization using techniques such as Hadamard incoherence, lattice-based codebooks, and learned additive quantization, often with lightweight fine-tuning.

\textbf{Weight Pruning.}
Weight pruning reduces model size by removing redundant parameters. Structured pruning, which eliminates entire architectural components (e.g., attention heads, MLP blocks), is preferred over unstructured pruning due to better hardware compatibility. Týr-the-Pruner~\cite{li2025t} introduces a global structure pruning framework based on search strategies to optimize sparsity patterns. AMP~\cite{mugnaini2025efficient} leverages input data projections to identify and prune less informative components in MHA and MLP layers. Wanda~\cite{sun2023simple} proposes a simple yet effective data-driven method that ranks weights for pruning based on their magnitudes and corresponding activation norms.

\section{Conclusion}
In this work, we propose FlexQuant, a dynamic token-wise quantization framework designed to bridge the gap between the memory demands of LLMs and modern hardware capacity. FlexQuant uses token-level perplexity entropy to guide fine-grained precision scheduling and applies KL-divergence-based layer-wise optimization to ensure global consistency. This dual-level strategy enables adaptive, efficient quantization with minimal accuracy loss, making FlexQuant well-suited for real-time LLM inference.

\section*{Acknowledgments}
This work is supported by the National Key Research and Development Program of China (2024YFE0204300), the National Natural Science Foundation of China (Grant No.62402311), and Natural Science Foundation of Shanghai (Grant No.24ZR1433700). This work is also sponsored by CAAI-Ant Group Research Fund.

\section*{Limitations}

Compared to PMPD using Any-Precision LLM to be able to recover arbitrarily low-precision weights from high-precision quantization weights, FlexQuant focuses on the quantization primitive, i.e., optimizing the access efficiency, and achieves arbitrary precision by storing multiple copies of the precision weights and KL-scattering-based inter-layer precision switching. Although some storage space is sacrificed, FlexQuant's weight switching is incremental and the switching order is determined offline. Therefore, models with different accuracies can be saved in inexpensive storage and prefetching the next round of switching to weights into the HBM is performed by GPU-Direct. Therefore, FlexQuant does not cause significant memory consumption and model loading blocking. In this paper, we only demonstrate the accuracy and inference performance of FlexQuant algorithm, models with different accuracy are loaded into HBM at the same time, and the subsequent need to implement the hard disk dynamic preloading mechanism in the system to adapt the real memory-constrained scenarios. 
\bibliography{custom}

\appendix

\end{document}